\documentclass[conference]{IEEEtran}

\usepackage[T1]{fontenc}  

\usepackage[utf8]{inputenc}
\usepackage{graphicx}
\usepackage{subcaption}

\RequirePackage[pdftex,hidelinks]{hyperref}
\usepackage{outlines}
\usepackage[shortlabels]{enumitem}
\usepackage[pdftex,dvipsnames,table,xcdraw]{xcolor}
\usepackage{colortbl}
\usepackage{graphicx}

\usepackage{booktabs}
\usepackage{multirow}
\usepackage{graphicx}
\usepackage{multirow}
\usepackage{multicol}
\usepackage{tabularx}

\usepackage{amsmath}

\begin{document}

\title{Evaluating GPT-4's Vision Capabilities on Brazilian University Admission Exams}

\author{\IEEEauthorblockN{Ramon Pires,
Thales Sales Almeida,
Hugo Abonizio, and
Rodrigo Nogueira}
\\
\IEEEauthorblockA{Maritaca AI\\
Brazil\\
}}

\maketitle

\begin{abstract}
Recent advancements in language models have showcased human-comparable performance in academic entrance exams.
However, existing studies often overlook questions that require the integration of visual comprehension, thus compromising the full spectrum and complexity inherent in real-world scenarios.
To address this gap, we present a comprehensive framework to evaluate language models on entrance exams, which incorporates both textual and visual elements. 
We evaluate the two most recent editions of \textit{Exame Nacional do Ensino Médio} (ENEM), the main standardized entrance examination adopted by Brazilian universities.
Our study not only reaffirms the capabilities of GPT-4 as the state of the art for handling complex multidisciplinary questions, but also pioneers in offering a realistic assessment of multimodal language models on Portuguese examinations.
One of the highlights is that text captions transcribing visual content outperform the direct use of images, suggesting that the vision model has room for improvement. 
Yet, despite improvements afforded by images or captions, mathematical questions remain a challenge for these state-of-the-art models. 
The code and data used on experiments are available at \url{https://github.com/piresramon/gpt-4-enem}.


\end{abstract}

\section{Introduction}
\label{sec:introduction}

Language models (LMs) have demonstrated remarkable performance on a variety of natural language processing tasks~\cite{devlin2018bert,brown2020language,rae2022scaling,chowdhery2022palm,hoffmann2022training,touvron2023llama}. 
Recent works have leveraged the capabilities of state-of-the-art LMs to evaluate automatic question resolution in college entrance exams~\cite{openai2023gpt4, zhang2023evaluating, nunes2023evaluating, almeida2023bluex}, shedding light on prospective advantages offered by LMs and their applicability in classroom settings \cite{kuvcak2018machine} and the evolving landscape of ``precision education'' \cite{luan2021review}. Furthermore, these models are increasingly applied across a diverse array of entrance exams spanning multidisciplinary domains, consistently exhibiting performance levels achieved by human students~\cite{openai2023gpt4, bommarito2022gpt, Bommarito2023Gpt, singhal2022large, nunes2023evaluating}.

However, many of these studies primarily focus on evaluating text-based models, often rejecting questions that necessitate image understanding.
Such approaches do not provide a realistic setting, limiting the potential richness and complexity of questions that demand a combination of both textual and visual comprehension. 

In this work, we address the diverse nature of entrance exams by incorporating image understanding.  We evaluate LMs on \textit{Exame Nacional do Ensino Médio} (\href{https://www.gov.br/inep/pt-br/areas-de-atuacao/avaliacao-e-exames-educacionais/enem}{\underline{ENEM}}), a multidisciplinary entrance exam for admission to high school education institutes in Brazil. The ENEM exam poses a unique challenge for LMs because it requires a deep comprehension across diverse fields of knowledge and the capability to synthesize information from varied domains.
We assess the two latest editions of ENEM: 2022 and 2023, with the latter applied and released to the public after the completion of the models' training.
Therefore, it is certain that the model did not encounter this particular exam during its training phase.
For all the experiments we apply few-shot with Chain-of-Thought (CoT)\cite{kojima2022large,wei2022chain} explanations, and evaluate the multimodal GPT-4~\cite{openai2023gpt4} as well as text-based language models. 

This study shows that language models can have impressive performance in solving multidisciplinary questions using step-by-step reasoning when applied in realistic settings with all the necessary visual and textual components.
However, the LMs exhibit relatively more difficulty in mathematics compared to other subjects. Furthermore, our findings suggest that captions prove more effective than images in current vision language models.

\begin{figure}[t]
\includegraphics[width=0.98\linewidth]{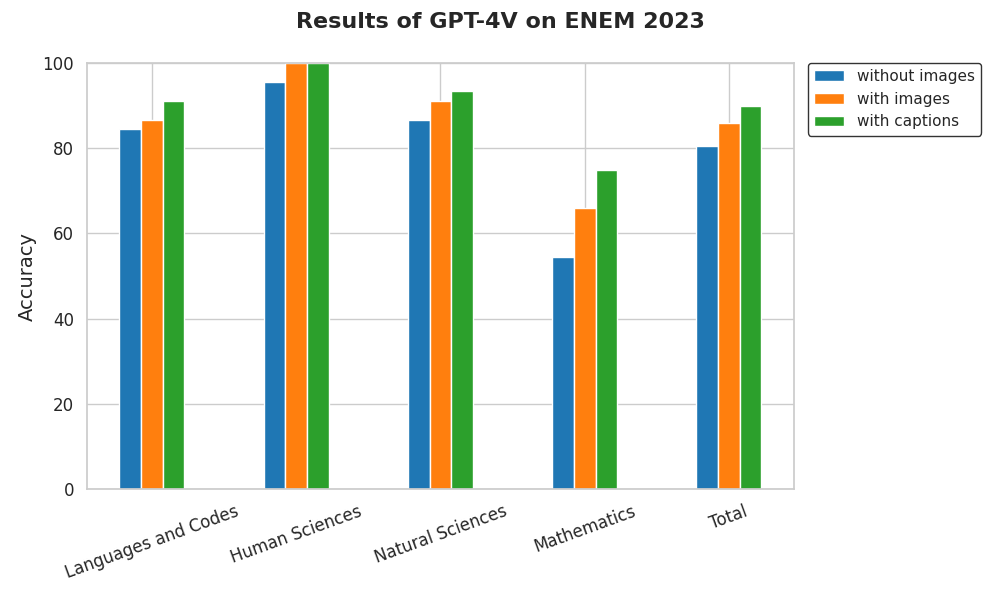} 
\caption{GPT-4V performance on the ENEM 2023 exam.}
\label{fig:plot_gpt4v}
\end{figure}

\section{Related Work}
\label{sec:related}

Some recent works have explored the capabilities of state-of-the-art language models in more challenging and domain-specific tasks~\cite{bommarito2022gpt, Bommarito2023Gpt, singhal2022large}.

GPT-4 exhibits human-level performance on multiple professional and academic exams, including multiple-choice and free-response questions, with images included in the prompt for questions that required them~\cite{openai2023gpt4}.
For all free-response questions, the images were objectively transcribed to evaluate vision and no-vision models.

Those benchmarks, however, are all English exams. Parallel to the growing interest in developing language models specialized in idioms~\cite{pires2023sabia}, it becomes urgent to propose benchmarks based on specific language examinations. Zhang et al.~\cite{zhang2023evaluating}, for example, evaluated the GPT-3.5 model on the Chinese college entrance examination (GAOKAO), excluding questions containing images.
Guillen-Grima et al.~\cite{guillengrima2023evaluating} assessed the performance of GPT-3.5 and GPT-4 models in the examination for access to Spanish Medical Residency. The authors evaluated the 2022 exam in the native language and English translations, and had a slightly enhanced performance with the last one. Similarly, the questions requiring image evaluations were excluded.

To bridge the gap of scarce benchmarks in Portuguese, Almeida et al.~\cite{almeida2023bluex} introduced the BLUEX (Brazilian Leading Universities Entrance eXams), a multimodal dataset consisting of the entrance exams of the two leading universities in Brazil: UNICAMP and USP. Despite being a multimodal benchmark, to date, the authors have solely evaluated text-based LMs on examples that do not require image comprehension.

Prior studies have evaluated the performance of NLP models on ENEM~\cite{silveira2018enem, nunes2023evaluating}. Silveira et al~\cite{silveira2018enem} employed static Word Embeddings~\cite{MikolovEfficientEstimationWord2013a} and WordNet~\cite{miller1995wordnet} to evaluate the ENEM from 2009 to 2017. The results were restricted, achieving an accuracy range of 26-29\%, even filtering out all the questions that require image understanding, mathematical reasoning, or having chemical elements, as well as the questions related to foreign languages.

More recently, Nunes et al.~\cite{nunes2023evaluating} evaluated text-based GPT-3.5 and GPT-4 models on ENEM 2009-2017 and ENEM 2022. The models showed satisfactory results and became capable of solving complex problems when decomposing them into smaller parts with CoT. However, they applied the same filtering criteria of~\cite{silveira2018enem}, eliminating nearly half of the questions.
We are the first to provide a realistic evaluation of multimodal language models on Portuguese benchmarks.

\section{Methodology}
\label{sec:methodology}

\subsection{Datasets}
\label{sec:datasets}


We evaluate the LMs using two datasets: ENEM 2022 and ENEM 2023. These datasets encompass all multiple-choice questions from the last two editions\footnote{Except the annulled questions. Each edition had one question of Mathematics annulled.}. The datasets have been created to allow the evaluation of both textual-only and textual-visual language models.
To evaluate textual-only models, we incorporated into the datasets the textual descriptions of the images that appear in the questions' statements from the orange ENEM exam booklet\footnote{A particular booklet that offers accessibility to people with visual impairments. The transcription of images, tables, and chemical equations are read by a reader during the exam.}. 

The datasets were created through a multi-step process. After downloading the booklets and respective answer sheets, both in PDF format, we use the Adobe PDFServices API~\cite{adobe-pdf-services-python-sdk} to extract text, tables, and images. The extraction process is followed by subsequent parsing procedures, and thorough manual verification and fixings. The last step was taken to remove accessibility patterns (e.g., replacing \textit{open quotation marks, sentence, close quotation marks} with \textit{"sentence"}) and converting mathematical expressions from natural language to standardized symbolic form (e.g., replacing \textit{v is equal to zero and a is different from zero} with \textit{v $=$ 0 and a $\neq$ 0}).


Different from the well-known datasets of ENEM exams~\cite{silveira2018enem,nunes2023evaluating}, our datasets address several key differences. First and foremost, our datasets keeps all the questions, including the ones that necessitate image understanding, table interpretation, mathematical reasoning, or familiarity with chemical elements. We also did not exclude the five questions related to the foreign language (English or Spanish, chosen at the time of registration) and kept the English ones. This comprehensive approach enables a more rigorous and nuanced evaluation process of language models across diverse questions.

In addition, a distinctive characteristic of our datasets lies in the inclusion of textual descriptions for visual elements, such as images and complex chemical equations. This method enables the evaluation of both multimodal and textual language models on the entrance exams.

Furthermore, our datasets incorporate table contents in Markdown formatting. This standardized structure is easily understandable by the language models, and because of this, we ignore the reader descriptions related to tables. 

The complete dataset comprises 358 multiple-choice multidisciplinary ENEM questions, encompassing all the textual, visual, and description information necessary to solve the exams. Table~\ref{tab:stats_enem} presents the statistics of the ENEM datasets, including the number of questions that require image understanding (IU).

\begin{table}[!tb]
\centering
\caption{Number of questions of ENEM 2022 and 2023 datasets.}
\begin{tabular}{@{}lcccc@{}}
\toprule
\multicolumn{1}{c}{\multirow{2}{*}{Area}} & \multicolumn{2}{c}{ENEM 2022} & \multicolumn{2}{c}{ENEM 2023} \\ \cmidrule(lr){2-3} \cmidrule(lr){4-5}
                      & total    & IU      & total    & IU      \\ \midrule
Languages and Codes   & 45       & 9       & 45       & 6       \\
Human Sciences        & 45       & 8       & 45       & 6       \\
Natural Sciences      & 45       & 15      & 45       & 13      \\
Mathematics           & 44       & 12      & 44       & 19      \\ \midrule
Total                 & 179      & 44      & 179      & 44      \\ \bottomrule
\end{tabular}
\label{tab:stats_enem}
\end{table}

\subsection{The OpenAI Models}

In this study, we use the following OpenAI's models available via API to paying customers\footnote{\url{https://platform.openai.com}}:

\subsubsection{\textbf{GPT-3.5 Turbo}} We use the most recent \texttt{gpt-3.5-turbo-1106}, whose training data is up to September 2021 and supports contexts with 16,385 tokens. 
\subsubsection{\textbf{GPT-4}} We use the latest version of GPT-4, \texttt{gpt-4-0613}, trained on data up to September 2021, enabling support for contexts with 8,192 tokens.
\subsubsection{\textbf{GPT-4 Turbo}} For comparative analysis, we employed the most recent \texttt{gpt-4-1106-preview}, trained on data up to April 2023, that can accommodate contexts with 128,000 tokens. This preview model is not currently recommended for production traffic.
\subsubsection{\textbf{GPT-4 Turbo with Vision}} We also extended the investigation to multimodal models, evaluating the latest version of GPT-4 with vision, the \texttt{gpt-4-vision-preview}. The model was trained on data up to April 2023, with support for contexts of 128,000 tokens. Similar to the GPT-4 Turbo, this preview model is not intended for production traffic at present. We will refer to this model as GPT-4V.

For all models, we used the following parameters: temperature of 0, which defaults to greedy decoding, maximum output of 512 tokens, frequency penalty of 0, and presence penalty of 0.

Since the last two models were trained with data up to April 2023 and the ENEM 2022 exam was applied on November 12-20 2022, we do not discard the possibility of data contamination. However, the results on the ENEM 2023 test are trustworthy, given that the models could not have access to the examination questions and answers during their training phase.

\subsection{Experiments}
\label{sec:experiments}

We adopted the prompt design that performed best in~\cite{nunes2023evaluating}: 3-shot with Chain-of-Thought (CoT)~\cite{DBLP:journals/corr/abs-2201-11903}. Those prompts include three few-shot examples from different knowledge areas of ENEM 2022, and include the solutions in explanatory steps, decomposing complex problems into smaller ones before compiling the final response. The Appendix~\ref{app:fewshot} presents the instruction, the few-shot examples and respective CoT explanations applied in the experiments.

We performed three experiments to analyze multiple-choice questions: without images, with images, and with captions. Each experiment is described hereafter.

\subsubsection{\textbf{Without Images}} In this experiment, we analyzed multiple-choice questions excluding any visual elements. This approach involved disregarding all images that could be present in both the question statement (and even in the answer alternatives). The primary aim was to lower-bound the language models' performance in the absence of visual contextual cues.


\subsubsection{\textbf{With Images}} For this experiment, we aim to evaluate multimodal LMs by embedding the images directly into the question body. Unlike the previous experiment, the inclusion of visual elements aimed to test the language models' ability to comprehend and process information from both textual and visual modalities.

\subsubsection{\textbf{With Captions}} In this experiment, our focus was on assessing textual LMs by replacing images with textual descriptions containing all essential information necessary for understanding the question. The objective was to measure the language models' proficiency in solving the exam when equipped with all the necessary information. This approach resembles LENS~\cite{berrios2023language}, a modular approach that fed LMs with image captions generated by vision models. The main difference herein is that we used human-generated captions.

\section{Results}
\label{sec:results}

This section presents the results for the two latest ENEM exams (2022 e 2023). All the experiments use 3-shot with CoT, that was the best approach in previous work~\cite{nunes2023evaluating}.

\subsection{Evaluating Multimodal Model}

\begin{table*}[t]
\centering
\caption{Results of GPT-4V on ENEM 2022 and ENEM 2023.}
\begin{tabular}{@{}lcccccc@{}}
\toprule
\multicolumn{1}{c}{\multirow{2}{*}{Area}} & \multicolumn{3}{c}{ENEM 2022} & \multicolumn{3}{c}{ENEM 2023} \\ \cmidrule(lr){2-4} \cmidrule(lr){5-7}
\multicolumn{1}{c}{} &
  \begin{tabular}[c]{@{}c@{}}without images\end{tabular} &
  \begin{tabular}[c]{@{}c@{}}with images\end{tabular} &
  \begin{tabular}[c]{@{}c@{}}with captions\end{tabular} &
  \begin{tabular}[c]{@{}c@{}}without images\end{tabular} &
  \begin{tabular}[c]{@{}c@{}}with images\end{tabular} &
  \begin{tabular}[c]{@{}c@{}}with captions\end{tabular} \\ \midrule
Languages and Codes	&	73.33	&	82.22	&	84.44	&	84.44	&	86.67	&	91.11	\\	
Human Sciences	    &	88.89	&	95.56	&	95.56	&	95.56	&	100.00	&	100.00	\\	
Natural Sciences	&	73.33	&	77.78	&	82.22	&	86.67	&	91.11	&	93.33	\\	
Mathematics     	&	54.55	&	61.36	&	61.36	&	54.55	&	65.91	&	75.00	\\	\midrule
Total           	&	72.63	&	79.33	&	81.01	&	80.45	&	86.03	&	89.94	\\ \bottomrule
\end{tabular}
\label{tab:results-gpt4-vision}
\end{table*}

Table~\ref{tab:results-gpt4-vision} shows the results of the OpenAI multimodal model -- GPT-4 Turbo with Vision --  over the two exams. 
We evaluate the model in the three experiments: without images, with images, and with captions. 
Both for ENEM 2022 and 2023, even without images, the GPT-4V model performed well in all the knowledge areas except in Mathematics (both reached 54.55\%). After including the images, the performance on ENEM 2022 in Mathematics improved to 61.36\% and remained the same when the images were replaced with captions. The ENEM 2023 had a remarkable improvement in the same area, achieving 65.91\% and 75.00\% with images and captions, respectively. 

The substantial improvement in Mathematics was expected given the model's initial difficulty in the without-images scenario. This enhancement is particularly noteworthy because Mathematics comprises a higher proportion of questions demanding image understanding (43\%).


Significant improvements are noticeable when incorporating either textual or visual representations of images, with the difference nearing 10 points, particularly when utilizing captions. This underscores the crucial role of having access to the complete statement. On ENEM 2023, noteworthy achievements are evident in Languages, Natural Sciences, and Human Sciences areas, all surpassing the 90\% accuracy, with special emphasis on the latter achieving a score of 100\%. Figure~\ref{fig:plot_gpt4v} illustrates the performance of the GPT-4V model on ENEM 2023.

The benefit of employing captions highlights that there is room for improvement in the encoding of images by vision models.

\subsection{Evaluating Textual-Only Models}

\begin{table*}[!htb]
\centering
\caption{Results of textual-only LMs on ENEM 2022.}
\begin{tabular}{@{}lcccccc@{}}
\toprule
\multicolumn{1}{c}{\multirow{2}{*}{Area}} & \multicolumn{2}{c}{GPT-3.5 Turbo} & \multicolumn{2}{c}{GPT-4} & \multicolumn{2}{c}{GPT-4 Turbo} \\ \cmidrule(lr){2-3} \cmidrule(lr){4-5} \cmidrule(lr){6-7}
                    & without images & with captions & without images & with captions & without images & with captions \\ \midrule
Languages and Codes &	68.89	&	77.78	&	75.56	&	88.89	&	73.33	&	86.67	\\
Human Sciences	    &	84.44	&	88.89	&	88.89	&	95.56	&	86.67	&	95.56	\\
Natural Sciences    &	71.11	&	66.67	&	84.44	&	86.67	&	77.78	&	80.00	\\
Mathematics 	    &	45.45	&	40.91	&	63.64	&	68.18	&	50.00	&	56.82	\\ \midrule
Total       	    &	67.60	&	68.72	&	78.21	&	84.92	&	72.07	&	79.89	\\ \bottomrule		
\end{tabular}
\label{tab:results-2022}
\end{table*}

\begin{table*}[!htb]
\centering
\caption{Results of textual-only LMs on ENEM 2023.}
\begin{tabular}{@{}lcccccc@{}}
\toprule
\multicolumn{1}{c}{\multirow{2}{*}{Area}} & \multicolumn{2}{c}{GPT-3.5 Turbo} & \multicolumn{2}{c}{GPT-4} & \multicolumn{2}{c}{GPT-4 Turbo} \\ \cmidrule(lr){2-3} \cmidrule(lr){4-5} \cmidrule(lr){6-7}
                    & without images & with captions & without images & with captions & without images & with captions \\ \midrule
Languages and Codes	&	80.00	&	82.22	&	84.44	&	93.33	&	84.44	&	91.11	\\
Human Sciences	    &	91.11	&	93.33	&	95.56	&	100.00	&	93.33	&	100.00	\\
Natural Sciences	&	75.56	&	73.33	&	88.89	&	95.56	&	86.67	&	86.67	\\
Mathematics     	&	38.64	&	45.45	&	65.91	&	72.73	&	59.09	&	65.91	\\ \midrule
Total           	&	71.51	&	73.74	&	83.80	&	90.50	&	81.01	&	86.03	\\ \bottomrule		
\end{tabular}
\label{tab:results-2023}
\end{table*}

As presented in the previous section, the use of captions is sufficient so that the model comprehends and solves the questions. Then, this section aims to evaluate other state-of-the-art language models that do not have the capability of understanding images. 
Tables~\ref{tab:results-2022} and ~\ref{tab:results-2023} show the results of GPT-3 Turbo, GPT-4, and GPT-4 Turbo when evaluating the multiple-choice questions of ENEM 2022 and 2023, respectively. 

Upon initial analysis of the results, intriguing patterns emerge. Firstly, even with the inclusion of captions, the GPT-3.5 Turbo had a performance decline in two areas for ENEM 2022 and one area for ENEM 2023, deviating from anticipated outcomes. Secondly, across all evaluated models, considerably superior results were achieved for ENEM 2023 compared to ENEM 2022, which raises the possibility that the most recent test had a lower level of difficulty.

Conversely, even though the GPT-4 Turbo was trained with more recent data (April 2023), it exhibited the smallest discrepancy in accuracy between ENEM 2023 (never seen during training) and ENEM 2022 (potentially seen during training). This observation challenges the contamination hypothesis, given that the model's performance was inferior to that of GPT-4.

Nevertheless, as expected, all models had a substantial improvement after including captions. The most notable advancement is evident in GPT-4, which had a 7-point enhancement, reaching 90.50\% with ENEM 2023, and surpassing the multimodal GPT-4V model.

\section{Conclusion}

This study extends the applicability of language models (LMs) by incorporating image understanding to address the diverse challenges posed by entrance exams.
Our evaluation reaffirms the GPT-4 as the state-of-the-art for tackling the ENEM, reinforcing findings from a previous study.
The multimodal GPT-4V LM exhibits notable performance across multiple experiments, but the superiority of the use of captions instead of images shows that there is room for improvement.
Through our analyses, we demonstrate the potential of LMs to contribute significantly to the educational landscape, which positions them as potential tools for enhancing educational assessments, offering perspectives on question difficulty prediction, generating new exam items, and ultimately helping the students in the preparation for admission exams

\section{Future Work}

For future work, we intend to include the essay in the evaluation and ask experts to correct the generated text using the same ENEM criteria. This would provide a more complete benchmark.

Additionally, we would like to adopt the same evaluation criteria currently adopted for the ENEM: the Item Response Theory (IRT). Different from the Classical Test Theory (CTC), which computes the total sum of correct answers regardless of successful guesses, the IRT is a mathematical model developed to make the test correction process fairer and more accurate.
This will allow a fair comparison of performance between LMs and students, checking the position of the models on the same ruler that measures the student's level of knowledge.

\section{Acknowledgments}
\label{sec:acks}
This research was partially funded by \textit{Fundação de Amparo à Pesquisa do Estado de São Paulo} (FAPESP), project 2022/01640-2.

\bibliographystyle{unsrt}
\bibliography{main.bib}

\clearpage

\appendix

\section{Appendices}

\subsection{Few-shot examples and CoT Explanations}
\label{app:fewshot}

This work uses the same few-shot examples and respective explanations used in~\cite{nunes2023evaluating}. The examples, as well as the instruction, are illustrated in Figure~\ref{fig:fewshot_examples_2022}.
The instruction and few-shot examples were translated into English to facilitate the understanding of the method, but we use it in Portuguese when evaluating the models.

\begin{figure}[!htb]
\sffamily
\tiny
\centering
\begin{tabular}{|
>{\columncolor[HTML]{D0F0C0}}p{0.9\columnwidth} |}
\hline
{\color[HTML]{343434} Formulate a step-by-step explanation that allows you to answer the multiple-choice question below. Only one alternative is correct. Desired format: point out the alternatives that make sense, choose the CORRECT alternative and justify it, and finish justifying why the other alternatives are incorrect. Finish the explanation with ``Answer:'' followed by the alternative.} \\
\hline
{\color[HTML]{343434} \textbf{Question 1:}}  \\
{\color[HTML]{343434} Emotional urgency. If everything is for yesterday, if life engages a first gear and takes off, if there is no more time for strategic stops, we fatally fall into the addiction of wanting love to be equally resolved in a split second. We are in a hurry to hear ``I love you". We can't wait for the rules of coexistence to be established: are we boyfriends, hookups, married, lovers? Emotional urgency. A trap. We associate several words with LOVE: passion, romance, sex, adrenaline, palpitation. We forget, however, the word that makes this feeling possible: ``patience". Love without patience does not avenge. Love cannot be chewed and swallowed with emergency, desperate hunger. It's a meal that can last a lifetime. MEDEIROS, M. Available at: http://porumavidasimples.blogspot.com.br. Accessed on: 20 Aug. 2017 (adapted).} \\
{\color[HTML]{343434} In this opinion text, the linguistic marks reveal a relaxed situation with little formality, which is evidenced by the} \\
{\color[HTML]{343434} A. impersonalization throughout the text, as in: ``if there is no more time''.} \\ 
{\color[HTML]{343434} B. construction of an atmosphere of urgency, in words like: ``hurry''.} \\ 
{\color[HTML]{343434} C. repetition of a certain syntactic structure, as in: ``If everything is for yesterday''.} \\ 
{\color[HTML]{343434} D. emphasis on the use of hyperbole, as in: ``a meal that can last a lifetime''.} \\ 
{\color[HTML]{343434} E. use of metaphors, as in: ``life engages a first gear and takes off''.} \\ 
{\color[HTML]{343434} \textbf{Explanation:} The text is written in a light, agile language, with little formality. In addition, it has figures of speech, such as metaphors and hyperboles, which are not mutually exclusive. In a sequential analysis of the alternatives, it would be possible to affirm that D) and E) are correct. However, looking in detail, it is noted that the expression ``use of metaphors'' proves to be more appropriate than ``emphasis on the use of hyperbole'', since, in order to state that the use of hyperbole was emphasized, the figure of speech should have appeared more often. This makes option E) more likely to be CORRECT. In addition, impersonality should not be pointed out as a mark of low formality. There is also an atmosphere of urgency, but that is criticized in the text that highlights the importance of patience and not haste. Finally, the syntactic structure is not systematically repeated throughout the text. \textbf{Answer:}} {\color[HTML]{00009B} \textit{E.}}     \\ \\
\hline \\
{\color[HTML]{343434} \textbf{Question 2:}}  \\
{\color[HTML]{343434} Whenever the relevance of discourse comes into play, the issue becomes political by definition, as it is the discourse that makes man a political being. And everything men do, know or experience only makes sense to the extent that it can be discussed. There will perhaps be truths that lie beyond language and that may be of great relevance to man in the singular, that is, to man who, whatever he may be, is not a political being. But men in the plural, that is, the men who live and move and act in this world, can only experience the meaning of things by being able to speak and be intelligible to each other and to themselves. ARENDT, H. The human condition. Rio de Janeiro: University Forensics, 2004.} \\
{\color[HTML]{343434} In the excerpt, the philosopher Hannah Arendt shows the importance of language in the process of} \\
{\color[HTML]{343434} A. understanding of culture.} \\ 
{\color[HTML]{343434} B. increased creativity.} \\ 
{\color[HTML]{343434} C. perception of individuality.} \\ 
{\color[HTML]{343434} D. improvement of technique.} \\ 
{\color[HTML]{343434} E. construction of sociability.} \\ 
{\color[HTML]{343434} \textbf{Explanation:} Hannah Arendt argues in her work that we are political beings, in the proper sense of living in a polis, in a collective and social environment. And this sociability is only possible through discourse, language. Thus, we can conclude that language is an important tool for building sociability, and therefore alternative E) is CORRECT. Furthermore, it is not about understanding the culture, but the social relationship between people of that culture. Hannah also doesn't talk about increased creativity, nor does she talk about technique. Finally, language is used in something more collective and social, just the opposite of individuality. \textbf{Answer:}} {\color[HTML]{00009B} \textit{E.}}     \\ \\
\hline \\
{\color[HTML]{343434} \textbf{Question 3:}}  \\
{\color[HTML]{343434} A couple plans to build a swimming pool in the shape of a rectangular parallelepiped with a capacity of 90,000 L of water on their farm. The couple hired a construction company that presented five projects with different combinations of the internal dimensions of depth, width and length. The pool to be built will have the same ceramic coating on its walls and bottom, and the couple will choose the project that requires the smallest coating area. The internal dimensions of depth, width and length, respectively, for each of the projects are: project I: 1.8 m, 2.0 m and 25.0 m; project II: 2.0 m, 5.0 m and 9.0 m; project III: 1.0 m, 6.0 m and 15.0 m; project IV: 1.5 m, 15.0 m and 4.0 m; project V: 2.5 m, 3.0 m and 12.0 m.} \\
{\color[HTML]{343434} The project that the couple should choose will be the} \\
{\color[HTML]{343434} A. I.} \\ 
{\color[HTML]{343434} B. II.} \\ 
{\color[HTML]{343434} C. III.} \\ 
{\color[HTML]{343434} D. IV.} \\ 
{\color[HTML]{343434} E. V.} \\ 
{\color[HTML]{343434} \textbf{Explanation:} We must calculate the area of the four side faces and the area of the lower base (bottom of the pool) and add these areas to obtain the coating area. Therefore, calculating the coating area of each project, we have: Project I: A = 2 x 25 + 2 x 1.8 x (2 + 25) = 147.2; Project II: A = 9 x 5 + 2 x 2 x (9 + 5) = 101; Project III: A = 15 x 6 + 2 x 1 x (15 + 6) = 132; Project IV: A = 4 x 15 + 2 x 1.5 x (15 + 4) = 117; Project V: A = 3 x 12 + 2 x 2.5 x (3 + 12) = 111. Therefore, the project with the smallest coating area is the project II, therefore the correct answer is B. \textbf{Answer:}} {\color[HTML]{00009B} \textit{B.}}     \\ \hline
\end{tabular}
\caption{Questions from the ENEM 2022 exam used as few-shot examples, and respective CoT explanations created from contents of high-school teachers' discussions.}
\label{fig:fewshot_examples_2022}
\end{figure}

\end{document}